\documentclass[A4paper, 11 pt, conference]{ieeeconf}
\IEEEoverridecommandlockouts       
\usepackage{geometry}
\usepackage{multirow}
\usepackage{amsmath}  
\usepackage{amssymb}  
\usepackage[colorlinks,urlcolor=blue,linkcolor=blue,citecolor=blue]{hyperref}
\usepackage{color,array}
\usepackage{graphicx}
\usepackage{cite}
\usepackage{setspace}  
\usepackage{subfigure}  
\usepackage{graphics} 
\usepackage{epsfig} 
\usepackage{times} 
\usepackage{amsmath} 
\usepackage{amssymb}  
\usepackage{subfigure}
\usepackage{algorithm} 
\usepackage{algorithmic} 
\usepackage{multirow} 
\usepackage{booktabs} 
\usepackage{gensymb}  
\usepackage{bm}  

\begin{document}

\title{\textbf{Mapping While Following: 2D LiDAR SLAM in Indoor Dynamic Environments with a Person Tracker}}

\author{Hanjing Ye$^1$, Guangcheng Chen$^1$, Weinan Chen$^2$, Li He$^2$, Yisheng Guan$^1$, and Hong Zhang$^{2*}$
\thanks{$*$ corresponding author (hzhang@sustech.edu.cn).}\thanks{$1$ Hanjing Ye, Guangcheng Chen and Yisheng Guan are with the Biomimetic and Intelligent Robotics Lab (BIRL), Guangdong University of Technology, Guangzhou, China, 510006.}\thanks{$2$ Hong Zhang, Weinan Chen, and Li He are with the Department of Electronic and Electrical Engineering, Southern University of Science and Technology, Shenzhen, China.}}

\maketitle
\thispagestyle{empty}

\begin{abstract}
    2D LiDAR SLAM (Simultaneous Localization and Mapping) is widely used in indoor environments due to its stability and flexibility. However, its mapping procedure is usually operated by a joystick in static environments, while indoor environments often are dynamic with moving objects such as people. The generated map with noisy points due to the dynamic objects is usually incomplete and distorted. To address this problem, we propose a framework of 2D-LiDAR-based SLAM without manual control that effectively excludes dynamic objects (people) and simplify the process for a robot to map an environment. The framework, which includes three parts: people tracking, filtering and following. We verify our proposed framework in experiments with two classic 2D-LiDAR-based SLAM algorithms in indoor environments. The results show that this framework is effective in handling dynamic objects and reducing the mapping error.
\end{abstract}

\section{INTRODUCTION}
2D LiDAR SLAM is a popular research topic in autonomous navigation\cite{cadena2016past}. It plays an important role in a number of applications with UGVs (Unmanned Ground Vehicles) operating in such environments as intelligent home, unmanned factory, shopping malls and so on. For most UGVs, the key step of 2D LiDAR SLAM is to integrate the current scan to the current map based on the odometry and the pose estimated by scan matching. In addition, it is generally assumed that the robot is moving in a static environment. However, real-world environments are usually populated with dynamic objects such as people. Scan points with noisy dynamic objects add uncertainty and thus can lead to inaccuracy of localization and mapping\cite{kim20191}\cite{pfreundschuh2021dynamic}.

Recently, many works\cite{underwood2013explicit}\cite{schauer2018peopleremover} about removing dynamic objects to build an accurate map purely from LiDAR point clouds have been proposed. With the development of LiDAR semantic segmentation\cite{milioto2019rangenet++, alonso20203d, zhu2021cylindrical}, points with class labels of dynamic objects can be directly excluded for improved localization and mapping\cite{zhao2019robust, ruchti2018mapping, sun2018recurrent}. However, these works are 3D-LiDAR-based with sufficient information so that it is possible to detect dynamic objects more easily than if a robot is equipped with only 2D-LiDAR.

\begin{figure}[t]
    \centering
    \includegraphics[width=0.45\textwidth]{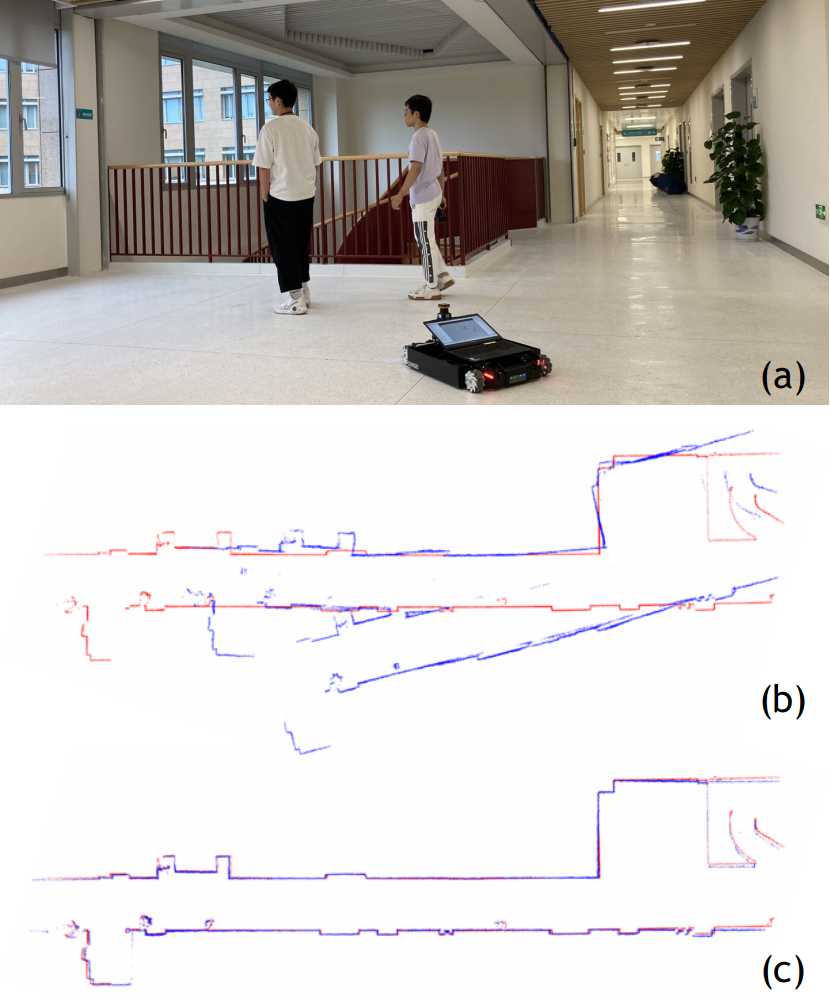}
    \caption{(a) People tracking and following in dynamic environments. With people following, the robot can generate a smooth trajectory with human-like motion. (b) Raw scan points with dynamic objects (people). With this noisy sensor input, the pose estimation of the robot is inaccurate, resulting a distorted map. (c) People are filtered from the scan points and, with this clean input, the generated map is more accurate and complete than (b).}
    \label{introduction}
\end{figure}

\begin{figure*}[t]
    \centering
    \includegraphics[width=0.80\textwidth]{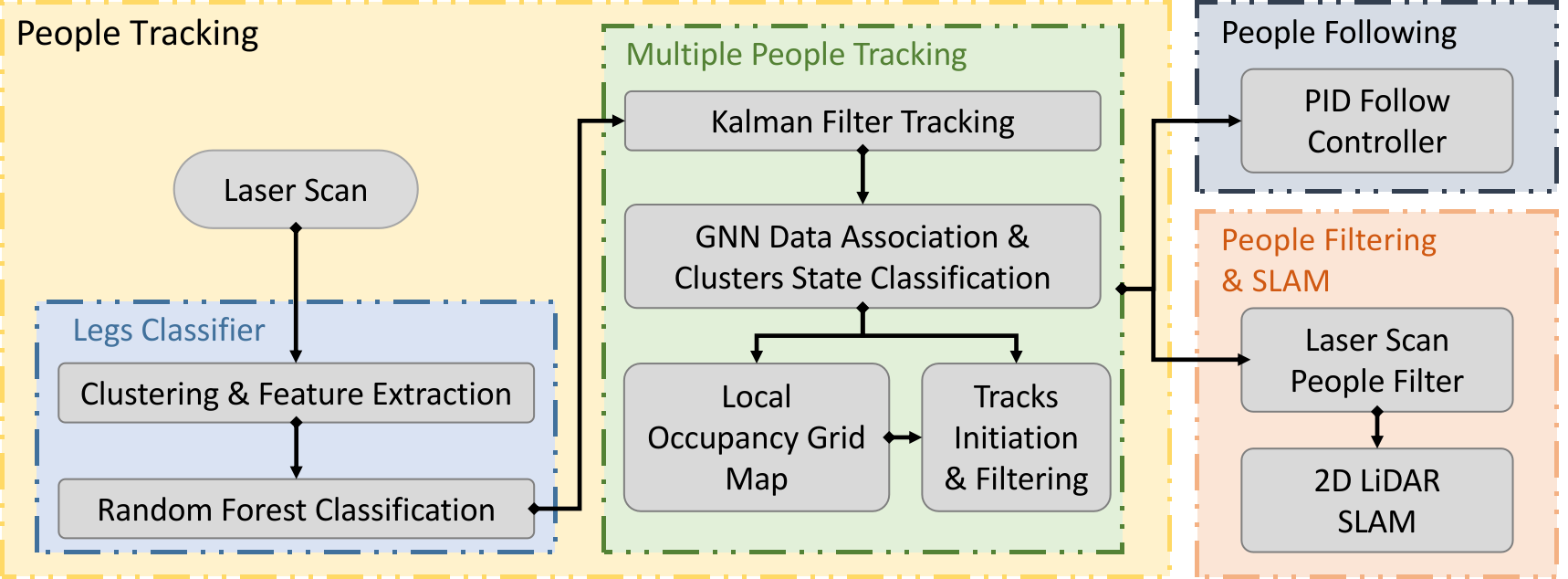}
    \caption{The framework comprising people tracking, filtering \& SLAM and following modules.}
    \label{method}
\end{figure*}
Most existing SLAM research assumes human intervention or interaction for navigation decision where a robot is manually driven, with a joystick for example. In other cases such as an assistive and service robot system\cite{cosgun2013autonomous}\cite{linder2016people}, a robot can make its navigation decision by tracking a person. In the SLAM procedure, in general, HRI\cite{goodrich2008human} can also provide a means for humans to convey their intentions to the robot, generate a purposeful robot motion, and improve mapping efficiency. In other words, people can directly guide a robot to explore an environment without a joystick. In this way, a human-like exploratory trajectory can be generated as an additional component in a complete SLAM system. However, a leading human represents a persistent dynamic object in front of the robot to interfere with its mapping operation, and the problem needs to be carefully managed, together with other dynamic objects, in order for the resulting map to be accurate. 

In this paper, we propose a framework to combine people tracking and following with dynamic object removal in 2D LiDAR SLAM. The people tracking method is based on our previous work\cite{leigh2015person}, which detects people by recognizing the two legs in a 2D LiDAR scan. We show through experiments that, with this framework, the interference due to people around the robot can be mitigated, resulting in a high-quality map. The ability to follow people while performing SLAM leads to a smooth motion trajectory and operational convenience.

\section{METHOD}

\subsection{People tracking}
The real-time people tracking algorithm\cite{leigh2015person} is employed in our work. We review the work in this section for the completeness of presentation. It comprises two steps: 1) LIDAR scan clustering and detection and 2) tracking with the combination of the Kalman filter\cite{kalman1960new} and GNN (Global Nearest Neighbor)\cite{konstantinova2003study} for data association.

Firstly, the points from LiDAR are clustered with a distance threshold $\alpha$. The threshold is small enough to separate two legs. Moreover, the clusters with less than $\beta$ points are considered as noise and discarded. The clusters are further classified for human recognition by a random forest classifier\cite{breiman2001random}.  Moreover, the random forest can output a confidence, and a threshold $\gamma$ is used to prune the false positives.
After the above LiDAR scan clustering and classification procedure, the detected clusters can be defined as:
\begin{equation}
\mathcal{\mathbf{Y}}_k = \{\mathbf{y}_{k}^{1}, \mathbf{y}_{k}^{2}, \mathbf{y}_{k}^{3}, \cdots, \mathbf{y}_{k}^{M_k}\}
\end{equation}
$\mathcal{\mathbf{Y}}_k$ represents the observation vector at timestep $k$ and it drives a Kalman filter to be described below. $M_k$ is the number of observations at time $k$. $\mathbf{y}_k^j = [x\ y]^T$ of a cluster is the position on a 2D plane.

Secondly, the positions of all the clusters are individually tracked using a Kalman filter. Data association of tracked clusters is solved by GNN. The set of tracked objects are defined as:
\begin{equation}
\mathcal{\mathbf{X}}_k = \{\mathbf{x}_{k}^{1}, \mathbf{x}_{k}^{2}, \mathbf{x}_{k}^{3}, \cdots, \mathbf{x}_{k}^{N_k}\}
\end{equation}
where $N_k$ is the number of tracked people at time $k$. State variables to be estimated for each person are defined as $\mathbf{x}_{k}^{j} = [x\ y\ \dot{x}\ \dot{y}]^T$ in terms of the position and the velocity of a cluster in 2D, $j = 1 ... N_k$. At timestep $k$, the motion model is defined as:
\begin{subequations}
\begin{align}
    \mathbf{x}_k^{j} = \mathbf{A}\mathbf{x}_{k-1}^{j} + \mathbf{w} \tag{3a}
\end{align}
where a constant velocity model is assumed, with the state transformation matrix:
\begin{align}
    \mathbf{A}=\begin{bmatrix}
        1&  0&  \Delta{t}& 0\\ 
        0&  1&  0& \Delta{t}\\ 
        0&  0&  1& 0\\ 
        0&  0&  0& 1
    \end{bmatrix} \tag{3b}
\end{align}
And the observation model is:
    \begin{align}
        \mathbf{y}_k^{j} = \mathbf{H}\mathbf{x}_{k}^{j} + \mathbf{v} \tag{3c}
    \end{align}
where observation only contains the position of a cluster, so the observation matrix is defined as:
\begin{align}
    \mathbf{H}=\begin{bmatrix}
        1&  0&  0& 0\\ 
        0&  1&  0& 0
    \end{bmatrix} \tag{3d}
\end{align}
\end{subequations}
The noises are assumed to be Gaussian white noise, where the covariance is defined as: $\mathbf{w}\sim\mathcal N(\mathbf{0}, \mathbf{Q})$ and $\mathbf{v}\sim\mathcal N(\mathbf{0}, \mathbf{R})$ where $\mathbf{Q}=q\mathbf{I}$ and $\mathbf{R}=r\mathbf{I}$. $q$ is dependent on the frequency of the LiDAR scan (smaller for higher frequency), and $r$ is determined by the accuracy of the LiDAR scan. A Kalman filter is then applied to update each person track $\mathbf{x}_{k}^{j}$:
\begin{subequations}
\begin{align*}
& \check{\mathbf{P}}_{k}^{j} = \mathbf{A}\hat{\mathbf{P}}_{k-1}^{j}\mathbf{A}^{T} + \mathbf{Q} \tag{4a}\\
& \check{\mathbf{x}}_{k}^{j} = \mathbf{A}\hat{\mathbf{x}}_{k-1}^{j} \tag{4b}\\
& \mathbf{K}_k^{j} = \check{\mathbf{P}}_{k}^{j}\mathbf{H}^{T}(\mathbf{H}\check{\mathbf{P}}_{k}^{j}\mathbf{H}^{T} + \mathbf{R})^{-1} \tag{4c}\\
& \hat{\mathbf{P}}_{k}^{j} = (\mathbf{I}-\mathbf{K}_{k}^{j}\mathbf{H})\check{\mathbf{P}}_{k}^{j} \tag{4d}\\
& \hat{\mathbf{x}}_k^{j} = \check{\mathbf{x}}_k^{j} + \mathbf{K}_k^{j}(\mathbf{y}_k^{j} - \mathbf{H}\check{\mathbf{x}}_k^{j}) \tag{4e}
\end{align*}
\end{subequations}
where the state initialization is defined as:
\begin{subequations}
\begin{align}
    \check{\mathbf{x}}_0^{j}=[x\ y\ 0\ 0]^T \tag{5a}
\end{align}
and the state covariance initialization is defined as:
\begin{align}
\check{\mathbf{P}}_0^{j}=\begin{bmatrix}
    0.05&  0&  0& 0\\ 
    0&  0.05&  0& 0\\ 
    0&  0&  0.05& 0\\ 
    0&  0&  0& 0.05
\end{bmatrix} \tag{5b}
\end{align}
\end{subequations}

To match observations $\mathbf{y}_k$ and tracks from the previous time $\mathbf{x}_{k-1}$ to update new tracks $\mathbf{x}_{k}$, the GNN data association algorithm of Munkres assignment \cite{kuhn1955hungarian} is used. With the Munkres assignment algorithm, a cost matrix in terms of the distance between detection $\mathbf{y}_k$ and previous track $\mathbf{x}_{k-1}$ is minimized. From the optimized cost matrix, an assignment between the detections and tracks can be obtained.

The previous tracks have two states: 1) the tracks $\mathbf{x}_{k-1}^j$ with matched detections $\mathbf{y}_k^j$, which can be updated with the Kalman filter to produce the new positions $\mathbf{x}_k^j$; 2) the tracks $\mathbf{x}_{k-1}^j$ without matched detections, which are updated with motion information only. The detections without matched tracks are initialized as new tracks with initial state of $\mathbf{x}_0^j=[x\ y\ 0\ 0]^T$. Every track maintains a confidence denoted as $c_k^j$, which is computed as following:
\begin{equation}
    c_k^j = 0.95c_{k-1}^j + 0.05d_c(y_k^i)
\end{equation}
where $d_c(y_k^i)$ is the confidence of $i_{th}$ detection of $j_{th}$ track. This confidence is produced by a random forest classifier that has been trained offline.

Furthermore, an odometry-corrected local occupancy grid map is built and updated for the detections without matched tracks. So all grid cells are defined to be \textit{free} if they are not occupied by non-human static objects detections.

To initialize the people tracks with stable performance, the following criteria are used for judging a pair of tracked clusters: 1) both are moved by a fixed distance $\delta$, 2) the confidences of both tracks are above a threshold $c_{min}$, and 3) both are in $free$ in the local occupancy grid map. Finally, when the first element of the state covariance $\check{\mathbf{P}}_{k}$ is larger than a threshold $\epsilon$ or if the confidence is below the threshold $c_{min}$, a people track is deleted.

\begin{table}[t]
    \caption{Comparison Among Three Methods for Guiding a Robot in Its Mapping Operation (the Number of $+$ Represents the Relative Performance)}
    \centering
    \scalebox{0.6}{
    \begin{tabular}{cccc}
        \toprule
        &\textbf{People following} &\textbf{Joystick control} &\textbf{Random exploration}\\
        \midrule
        Convenience &$++$ &$+$ &$+++$\\
        Time saving &$+++$ &$+++$ &$+$\\
        Motion smoothness &$+++$ &$+$ &$++$\\
        Safety &$+++$ &$++$ &$+$\\
        \bottomrule
    \end{tabular}}
    \label{people following comparison}
\end{table} 

\subsection{People filter and SLAM}
After people detection initialization and deletion, stable people tracks can be obtained. So in the local occupancy grid map, there are two kinds of occupied grid cells in terms of people and non-people. Further, since the reference frame of the local occupancy grid map is the same as the raw scan, we can filter the dynamic objects directly with the position estimation of people tracks.
Specifically, for each LiDAR scan point, it is considered as coming from a person if its Euclidean distance from any of the people tracks is less than a given threshold $\zeta$. A larger threshold indicates fewer false positives of dynamic objects (i.e., people), at the expense of a less complete scan due to wrong deletion of the static objects.

Once points in the LiDAR scan that belong to people have been removed, the filtered scan can be fed into 2D LiDAR SLAM. Our framework works for any such algorithms but we examine two popular representatives in our research: GMapping\cite{grisetti2007improved} and Cartographer\cite{hess2016real}. GMapping is a filter-based SLAM, where wheel odometry is used for prediction, and the estimated pose by scan-map matching for correction. Cartographer is an optimization-based SLAM with well-designed filters and optimization processes. Given odometry information and LIDAR scan, Cartographer firstly uses voxel filters to down-sample the scan. After that, optimization-based scan-map matching is used to add a new scan to its current submap while the wheel odometry information is used as an initialized pose. Lastly, global optimization is run to align the submaps for a coherent global map. In our experiments, to be described shortly, we use non-filtered and filtered scans as input separately to demonstrate the benefit of dynamic object filtering in terms of its adaptation ability.

\subsection{People following}
Among the methods used in guiding a robot in its mapping procedure, our comparison between people following, joystick control and random exploration is shown in Table \ref{people following comparison}. It shows that guiding robot mapping by people following could potentially have significant advantages.

At the onset of each experiment, a target person, for example, a detected person closest to the robot, is selected to be followed. At each time instance, the position error is computed from the followed person with respect to the robot. With the aim of maintaining a fixed following distance, a trajectory toward the followed person can be generated. For the generation of a smooth and accurate following trajectory, a PID (Proportional-Integral-Derivative) controller is used for motion execution. In our control strategy, only angular and linear velocity of the robot are controlled independently, although our robot in this study (Clearpath Dingo) is holonomic.

\section{EXPERIMENTS AND RESULTS}

\subsection{Environmental setup}
\begin{figure}[t]
    \centering
    \subfigure[corridor]{
        \centering
        \includegraphics[width=0.22\textwidth]{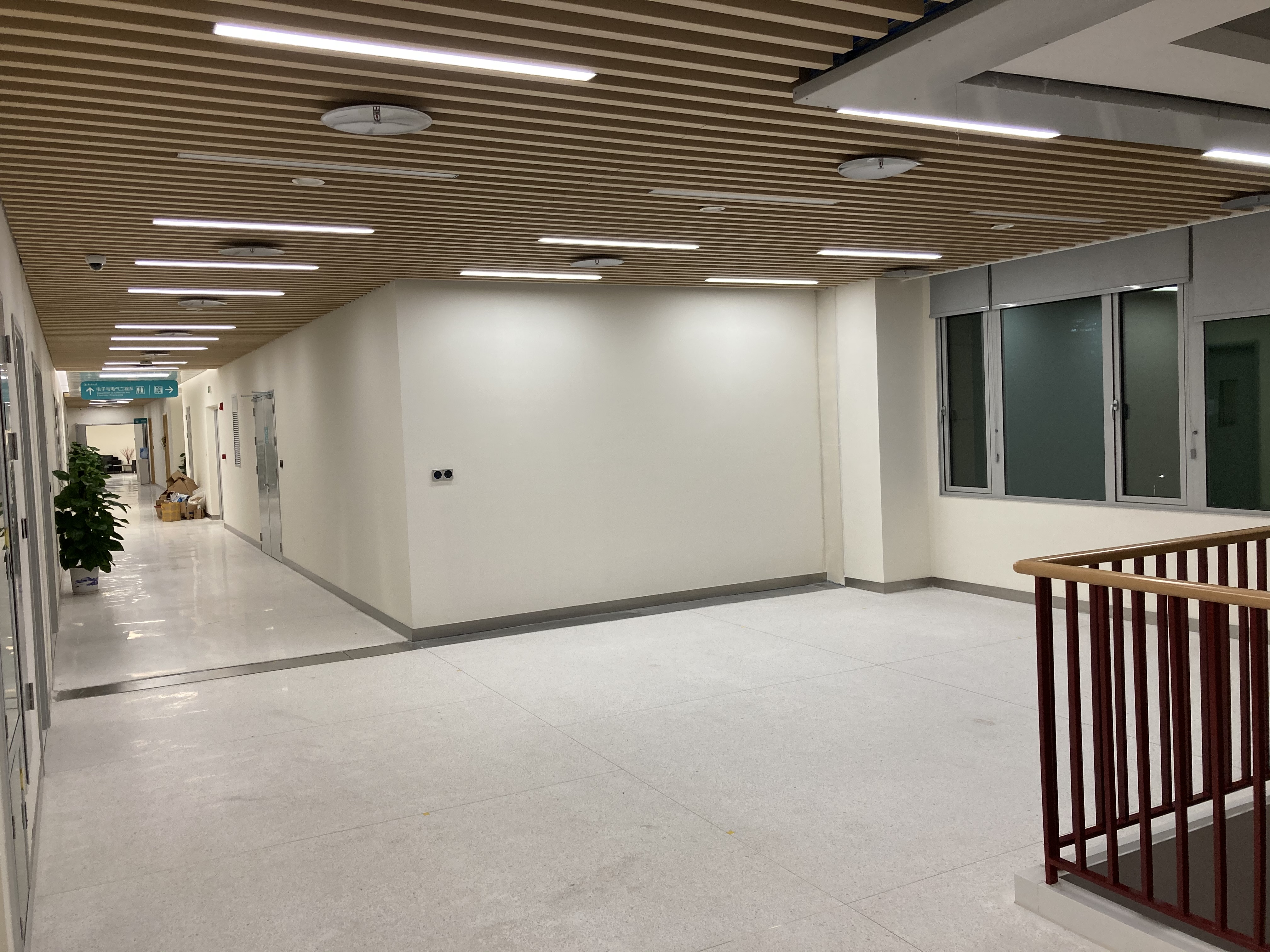}
    }
    \subfigure[meeting room]{
        \centering
        \includegraphics[width=0.22\textwidth]{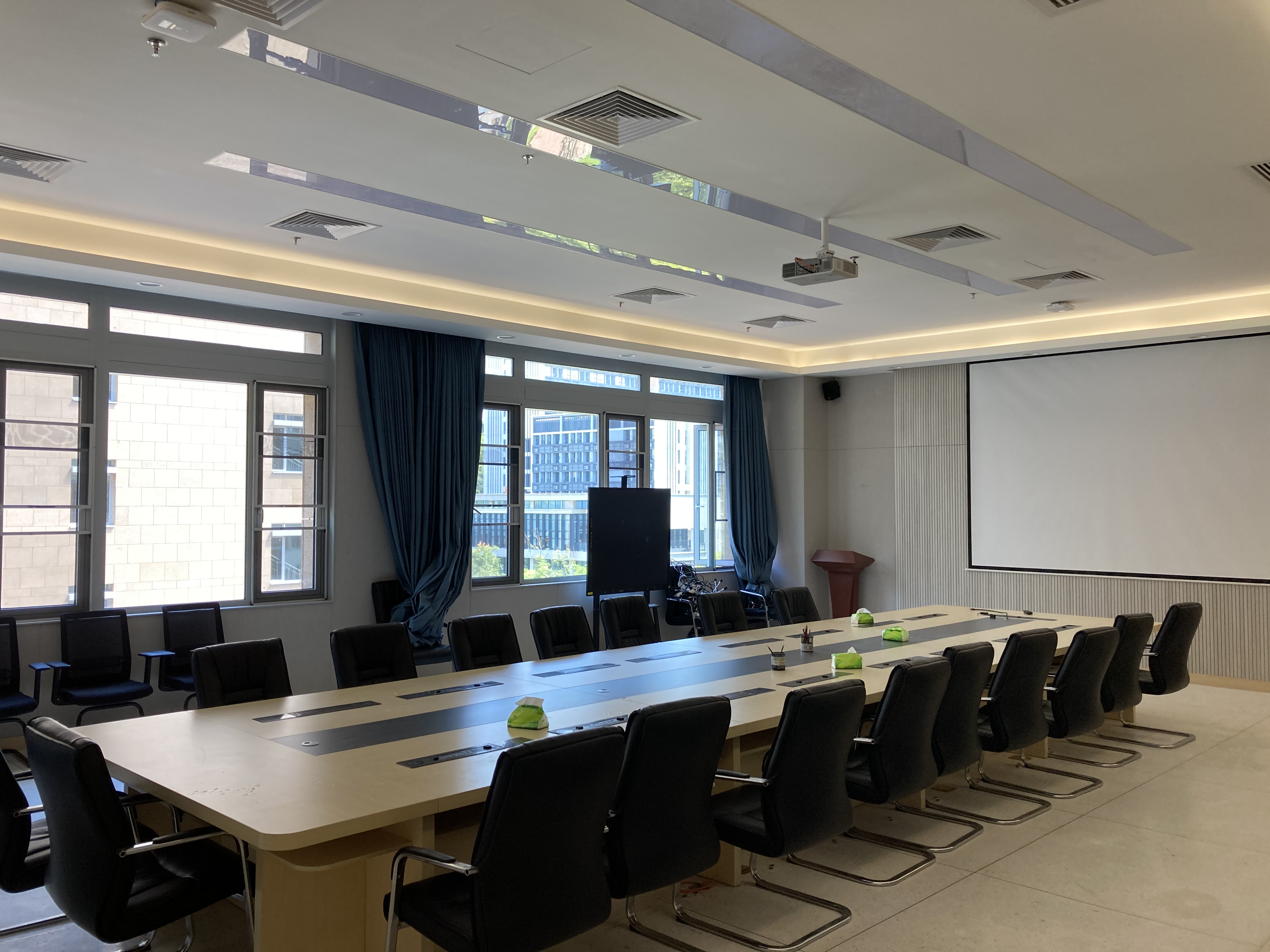}
    }
    \caption{The environments with a) a long corridor with a platform; b) a meeting room with chairs and a big table.}
    \label{environment}
\end{figure}

We evaluate our framework with two LIDAR SLAM algorithms (GMapping\cite{grisetti2007improved} and Cartographer\cite{hess2016real}) in two environments including a corridor and a meeting room in a typical office building, shown in Figure \ref{environment}. Each experiment is composed of three sub-experiments including \textit{no-people, 1-person and 2-people}. \textit{No-people} means there are no people during the mapping procedure, and this experiment involves a static environment, and produces the baseline performance. It is worth mentioning that the no-people experiment is carefully conducted with slow and smooth motion so that the generated map is of high-quality enough to be used as our ground truth map. \textit{1-person} implies the people following is added with one leading person, and \textit{2-people} is conducted with two people walking in front of the robot, one of whom is the leader to be tracked. Each experiment of \textit{1-person} and \textit{2-people} is repeated 5 times to allow us to draw statistical conclusions.

Some detailed implementations are as follows. All the thresholds in people tracking, shown in Table \ref{threshold}, can be adjusted according to the actual experiment. The resolution of the generated map is set to 0.02 m for both GMapping and Cartographer. The experiment is conducted based on ROS (Robot Operating System). A Clearpath Dingo-O, a Hokuyo UTM-30LX and a laptop with Intel(R) Core(TM) i5-10200H CPU @ 2.40GHz are used in the experiments. For the Hokuyo UTM-30LX, its detection range is 0.1-30 m in a field of view of 270{\degree} at the angular resolution of 0.25\degree. The scanning frequency is 40 Hz.

\begin{table}[t]
    \caption{Distance threshold (m) and other scalars. \bm{$\alpha$} indicates the minimum distance of leg point. \bm{$\beta$} is the minimum number of points for clustering. \bm{$\gamma$} is a confidence threshold for reducing false positives of clusters. \bm{$\delta$} is a distance threshold for tracks initiation. \bm{$\epsilon$} is the minimum state covariance for deleting wrong tracks. \bm{$\zeta$} is a distance threshold for people filtering. \bm{$c_{min}$} is a confidence threshold for people deletion. \bm{$q$} and \bm{$r$} are covariance factor for Kalman filter.}
    \centering
    \scalebox{0.9}{
    \begin{tabular}{ccccccccc}
        \toprule
        \bm{$\alpha$} &\bm{$\beta$} &\bm{$\gamma$} &\bm{$\delta$} &\bm{$\epsilon$} &\bm{$\zeta$} &\bm{$c_{min}$} &\bm{$q$} &\bm{$r$} \\
        \midrule
        0.13 &3 &0.3 &0.5 &0.9 &0.35 &0.1 &0.05 &0.1\\
        \bottomrule
    \end{tabular}}
    \label{threshold}
\end{table} 

\subsection{Evaluation metric}
In this paper, ADNN\cite{santos2013evaluation} (average distance to the nearest neighbor) is calculated to characterize the quality of the map built, and it serves as the performance metric to evaluate different algorithms.
ADNN is calculated after first registering the estimated map and ground truth (the map generated with \textit{no-people} setting). We adopt PCA\cite{jolliffe2016principal} to align their centroids and the angle between their first eigenvectors for coarse registration, followed by an ICP\cite{pomerleau2015review} for fine registration. Secondly, NN (nearest neighbor)$\footnote{A function for finding the nearest point and returning an euclidean distance from Scikit-learn\cite{pedregosa2011scikit}}$ search is used to find the correspondences between points in the map being evaluated and those in the ground truth map. ADNN is calculated as below:
\begin{equation}
ADNN = \frac{1}{N}\sum_{i=1}^{N}\sqrt{(u_{i}^{est}-u_{i}^{gt})^2+(v_{i}^{est}-v_{i}^{gt})^2}
\end{equation}
where $N$ is the number of the occupied cells in the ground truth map. $u$ and $v$ are the coordinates of the map, and point at $(u_{i}^{est}, v_{i}^{est})$ in the estimated map is the nearest neighbor of point $(u_{i}^{gt}, v_{i}^{gt})$ in the ground truth map.

\subsection{Results and discussion}

\begin{figure}[t]
    \centering
    \subfigure[GMapping]{
        \centering
        \includegraphics[width=0.45\textwidth]{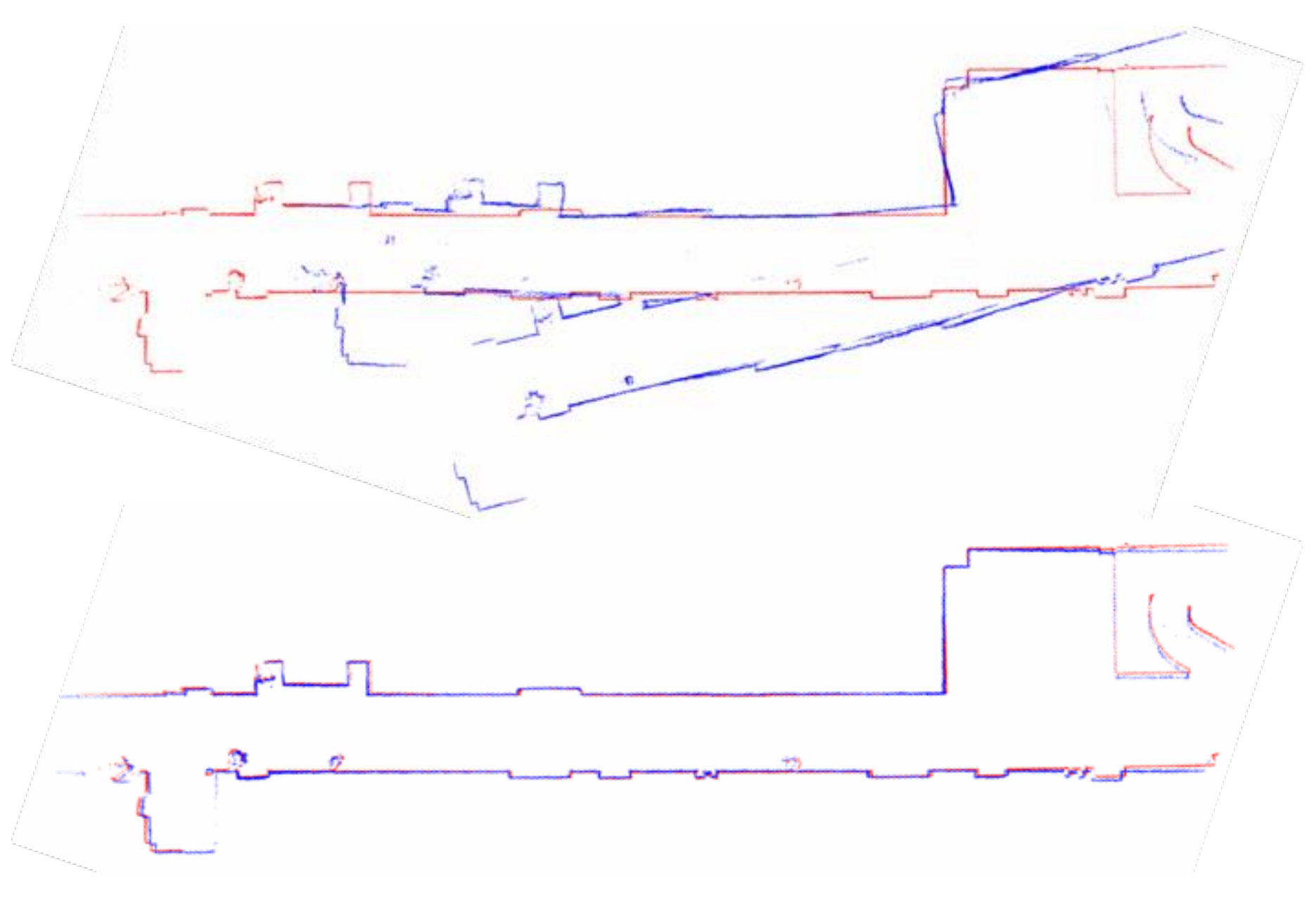}
        \label{maps_corridor_gmapping}
    }
    \subfigure[Cartographer]{
        \centering
        \includegraphics[width=0.45\textwidth]{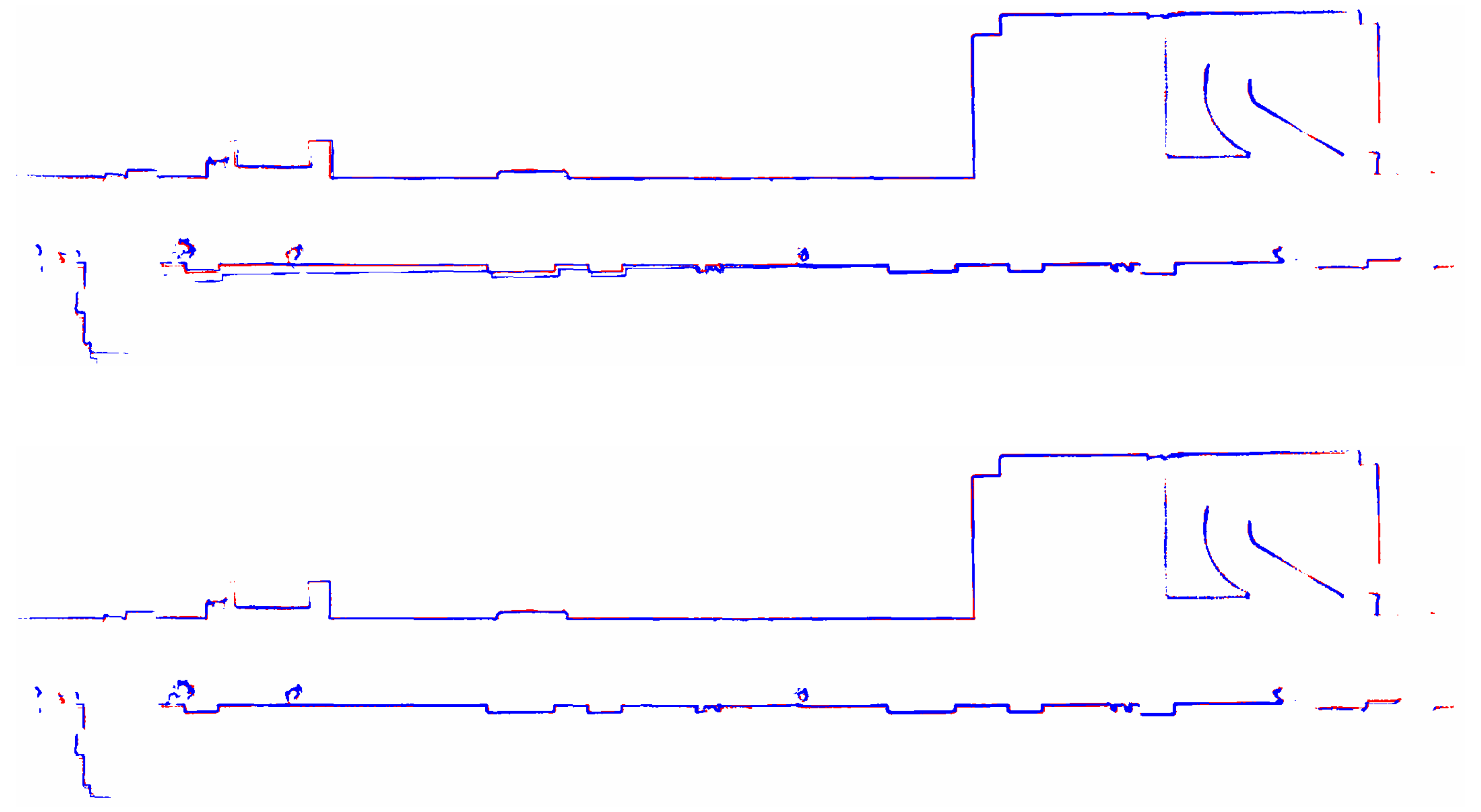}
        \label{maps_corridor_car}
    }
    \caption{Generated maps of the corridor. The red one is the ground truth map and the blue one is the estimated map with the \textit{2-people} setting. The upper one uses raw LIDAR scans for mapping and the bottom one uses filtered scans with dynamic object removal are as input.}
    \label{maps_corridor}
\end{figure}

\begin{figure}[ht]
    \centering
    \subfigure[GMapping]{
        \centering
        \includegraphics[width=0.22\textwidth]{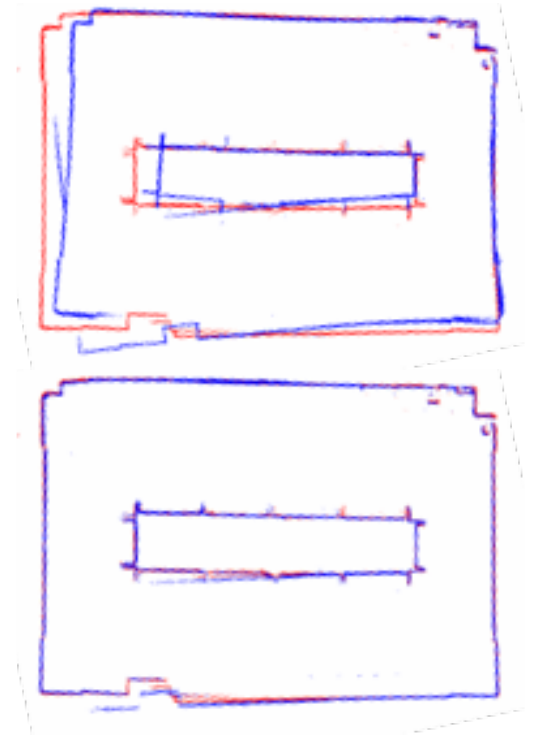}
        \label{maps_meetingRoom_gmapping}
    }
    \subfigure[Cartographer]{
        \centering
        \includegraphics[width=0.22\textwidth]{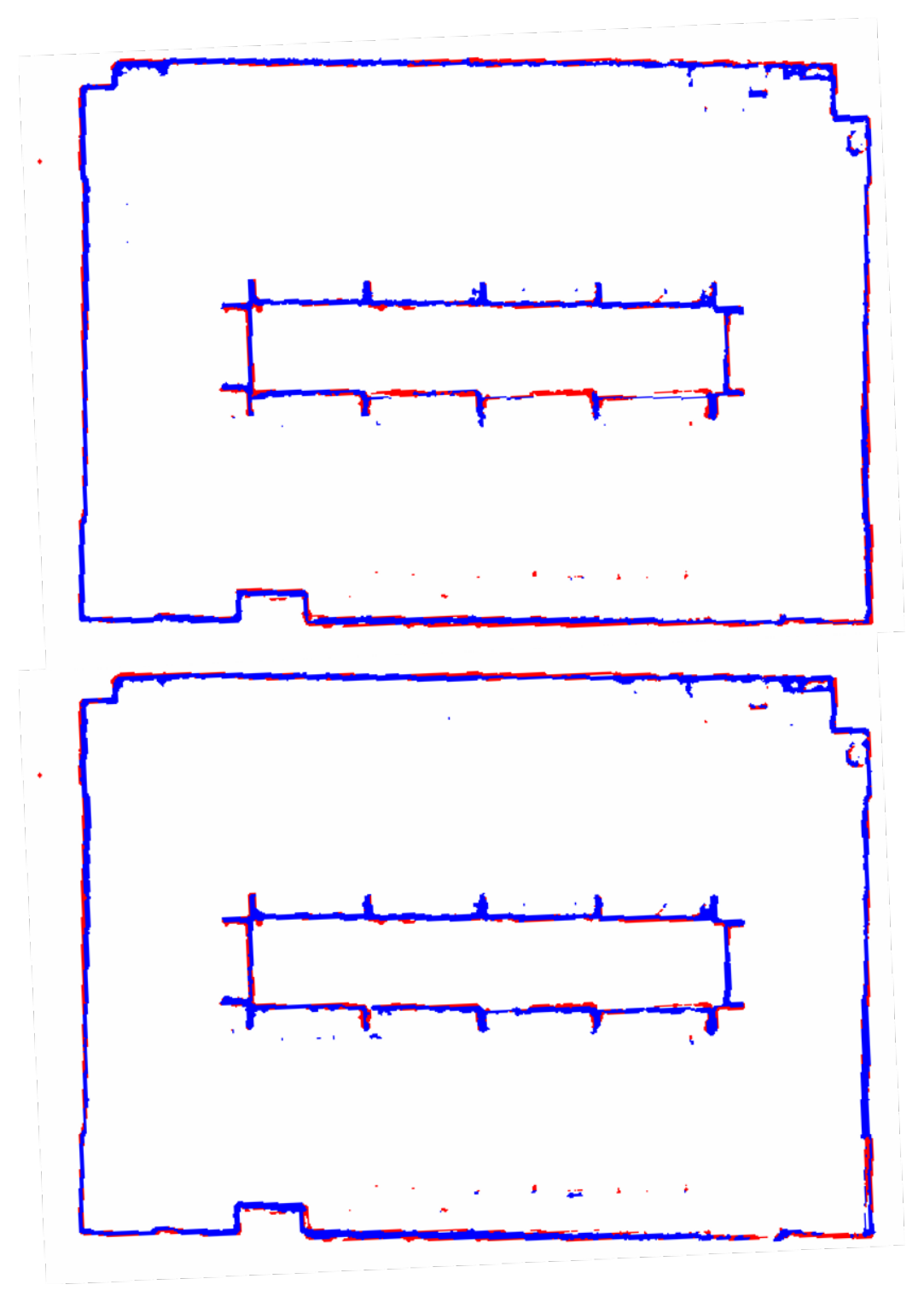}
        \label{maps_meetingRoom_car}
    }
    \caption{Generated maps of the meeting room.}
    \label{maps_meetingRoom}
\end{figure}

\begin{figure*}[t]
    \centering
    \subfigure[Corridor]{
        \centering
        \includegraphics[width=0.80\textwidth]{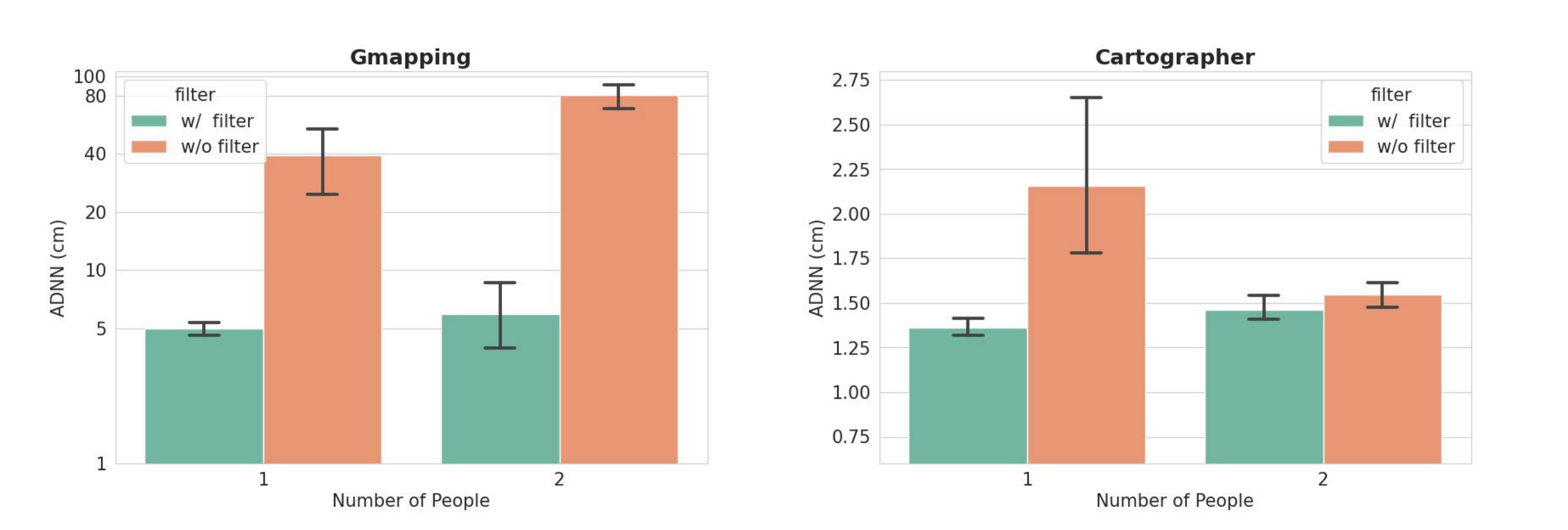}
        \label{plots_corridor}
    }
    \subfigure[Meeting room]{
        \centering
        \includegraphics[width=0.80\textwidth]{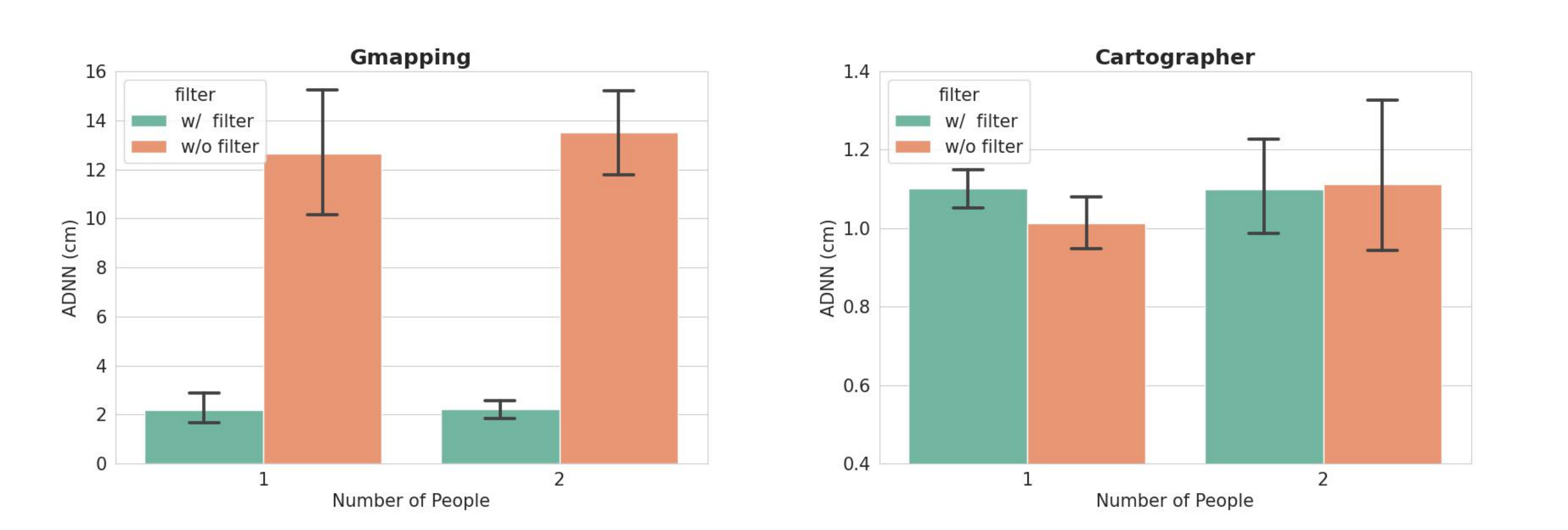}
        \label{plots_meetingRoom}
    }
    \caption{Box plots of mapping error with/without dynamic object removal. (a) Experiments are conducted in the corridor, where the mean error with filtered scan is 5.90 cm versus 80.24 cm of the non-filtered scan in the case of \textit{2-people} using GMapping. (b) In the meeting room, dynamic object removal is also beneficial for GMapping, while it is not improved much for Cartographer.}
    \label{plots}
\end{figure*}

For the results of the corridor, as shown in Figure \ref{maps_corridor_gmapping}, the map with raw scans is with serious distortion, while the map with filtered ones is almost the same as the ground truth map. Quantitatively, as shown in the GMapping plot of the Figure \ref{plots_corridor}, for all metrics of the mapping quality, e.g., mean, maximum error and  minimum error, the filtered case with dynamic object removal is superior to the one without filtering. Specifically, in the \textit{1-person}, the mean error of our algorithm with scan filtering is 4.98 cm versus 39.36 cm of the non-filtered case. While in the \textit{2-people} experiment, it is 5.90 cm versus 80.24 cm. For Cartographer, as shown in Figure \ref{maps_corridor_car} and \ref{plots_corridor}, the mean error of the filtered one is 1.36 cm lower than 2.16 cm of the non-filtered one. Although the difference of mean ADNN is minor, the edge of the map built with non-filtered scans is clearly flawed with two linear structures for the same wall. Besides, even with only raw scans without filtering, Cartographer performs better than GMapping significantly.

The superior performance of Cartographer to GMapping in the case of dynamic objects could be attributed to their internal algorithmic steps. GMapping is a filter-based SLAM without explicit steps for removing outliers. Therefore a scan with dynamic objects can easily impact the performance of the system. In contrast, Cartographer is an optimization-based SLAM and its voxel filter is effective in mitigating the effect of the noisy LiDAR points due to dynamic objects. The points of the legs could be sparser with voxel filter down-sampling in Cartographer, and local and global optimizations can minimize the error introduced by scan points due to dynamic objects.

For the performance in the meeting room, as shown in Figures \ref{maps_meetingRoom} and \ref{plots_meetingRoom}, GMapping can perform better with filtered scans, reducing the error to 10.46 cm mean error in the \textit{1-person} case and to 10.30 cm in the \textit{2-people} case. Cartographer acts similarly with filtered scans and raw scans. The similar performance of Cartographer with and without filtering in this environment could be due to the fact that the meeting room is smaller than the corridor and full of distinctive physical features. So the optimization-based scan matching can provide more precise pose estimation. However, comparing both generated maps, the map with filtered scans is more complete while the map with raw scans lack some edges of the central table.

From the above observations and analysis, we can infer that, in the experimental environments, the effect of the people can be efficiently mitigated with our framework.

\section{CONCLUSION}
We propose a framework comprising people tracking, following and filtering for 2D LiDAR SLAM in indoor dynamic environments. The results of the experiments indicate that the framework is effective in removing dynamic objects and improving map accuracy in dynamic environments. Another observation is that Cartographer is more robust to dynamic people than GMapping. Lastly, if people following is a convenient way to guide a robot through an environment and build its map, then our framework provides a solution that effectively removes the effect of interference caused by people moving in front of the robot. 

In the future, several works can be expanded from this framework, including evaluating the framework in various environments with a large range of complexity in terms of the number of people and combining vision with LiDAR in detecting, tracking and filtering dynamic objects.

\bibliographystyle{IEEEtran}
\bibliography{ref}

\begin{thebibliography}{10}
\providecommand{\url}[1]{#1}
\csname url@samestyle\endcsname
\providecommand{\newblock}{\relax}
\providecommand{\bibinfo}[2]{#2}
\providecommand{\BIBentrySTDinterwordspacing}{\spaceskip=0pt\relax}
\providecommand{\BIBentryALTinterwordstretchfactor}{4}
\providecommand{\BIBentryALTinterwordspacing}{\spaceskip=\fontdimen2\font plus
\BIBentryALTinterwordstretchfactor\fontdimen3\font minus
  \fontdimen4\font\relax}
\providecommand{\BIBforeignlanguage}[2]{{%
\expandafter\ifx\csname l@#1\endcsname\relax
\typeout{** WARNING: IEEEtran.bst: No hyphenation pattern has been}%
\typeout{** loaded for the language `#1'. Using the pattern for}%
\typeout{** the default language instead.}%
\else
\language=\csname l@#1\endcsname
\fi
#2}}
\providecommand{\BIBdecl}{\relax}
\BIBdecl

\bibitem{cadena2016past}
C.~Cadena, L.~Carlone, H.~Carrillo, Y.~Latif, D.~Scaramuzza, J.~Neira, I.~Reid,
  and J.~J. Leonard, ``Past, present, and future of simultaneous localization
  and mapping: Toward the robust-perception age,'' \emph{IEEE Transactions on
  robotics}, vol.~32, no.~6, pp. 1309--1332, 2016.

\bibitem{kim20191}
G.~Kim, B.~Park, and A.~Kim, ``1-day learning, 1-year localization: Long-term
  lidar localization using scan context image,'' \emph{IEEE Robotics and
  Automation Letters}, vol.~4, no.~2, pp. 1948--1955, 2019.

\bibitem{pfreundschuh2021dynamic}
P.~Pfreundschuh, H.~F.~C. Hendrikx, V.~Reijgwart, R.~Dub{\'e}, R.~Siegwart, and
  A.~Cramariuc, ``Dynamic object aware lidar slam based on automatic generation
  of training data,'' \emph{arXiv preprint arXiv:2104.03657}, 2021.

\bibitem{underwood2013explicit}
J.~P. Underwood, D.~Gillsj{\"o}, T.~Bailey, and V.~Vlaskine, ``Explicit 3d
  change detection using ray-tracing in spherical coordinates,'' in \emph{2013
  IEEE international conference on robotics and automation}.\hskip 1em plus
  0.5em minus 0.4em\relax IEEE, 2013, pp. 4735--4741.

\bibitem{schauer2018peopleremover}
J.~Schauer and A.~N{\"u}chter, ``The peopleremover—removing dynamic objects
  from 3-d point cloud data by traversing a voxel occupancy grid,'' \emph{IEEE
  robotics and automation letters}, vol.~3, no.~3, pp. 1679--1686, 2018.

\bibitem{milioto2019rangenet++}
A.~Milioto, I.~Vizzo, J.~Behley, and C.~Stachniss, ``Rangenet++: Fast and
  accurate lidar semantic segmentation,'' in \emph{2019 IEEE/RSJ International
  Conference on Intelligent Robots and Systems (IROS)}.\hskip 1em plus 0.5em
  minus 0.4em\relax IEEE, 2019, pp. 4213--4220.

\bibitem{alonso20203d}
I.~Alonso, L.~Riazuelo, L.~Montesano, and A.~C. Murillo, ``3d-mininet: Learning
  a 2d representation from point clouds for fast and efficient 3d lidar
  semantic segmentation,'' \emph{IEEE Robotics and Automation Letters}, vol.~5,
  no.~4, pp. 5432--5439, 2020.

\bibitem{zhu2021cylindrical}
X.~Zhu, H.~Zhou, T.~Wang, F.~Hong, Y.~Ma, W.~Li, H.~Li, and D.~Lin,
  ``Cylindrical and asymmetrical 3d convolution networks for lidar
  segmentation,'' in \emph{Proceedings of the IEEE/CVF Conference on Computer
  Vision and Pattern Recognition}, 2021, pp. 9939--9948.

\bibitem{zhao2019robust}
S.~Zhao, Z.~Fang, H.~Li, and S.~Scherer, ``A robust laser-inertial odometry and
  mapping method for large-scale highway environments,'' in \emph{2019 IEEE/RSJ
  International Conference on Intelligent Robots and Systems (IROS)}.\hskip 1em
  plus 0.5em minus 0.4em\relax IEEE, 2019, pp. 1285--1292.

\bibitem{ruchti2018mapping}
P.~Ruchti and W.~Burgard, ``Mapping with dynamic-object probabilities
  calculated from single 3d range scans,'' in \emph{2018 IEEE International
  Conference on Robotics and Automation (ICRA)}.\hskip 1em plus 0.5em minus
  0.4em\relax IEEE, 2018, pp. 6331--6336.

\bibitem{sun2018recurrent}
L.~Sun, Z.~Yan, A.~Zaganidis, C.~Zhao, and T.~Duckett, ``Recurrent-octomap:
  Learning state-based map refinement for long-term semantic mapping with
  3-d-lidar data,'' \emph{IEEE Robotics and Automation Letters}, vol.~3, no.~4,
  pp. 3749--3756, 2018.

\bibitem{cosgun2013autonomous}
A.~Cosgun, D.~A. Florencio, and H.~I. Christensen, ``Autonomous person
  following for telepresence robots,'' in \emph{2013 IEEE International
  Conference on Robotics and Automation}.\hskip 1em plus 0.5em minus
  0.4em\relax IEEE, 2013, pp. 4335--4342.

\bibitem{linder2016people}
T.~Linder and K.~O. Arras, ``People detection, tracking and visualization using
  ros on a mobile service robot,'' in \emph{Robot Operating System
  (ROS)}.\hskip 1em plus 0.5em minus 0.4em\relax Springer, 2016, pp. 187--213.

\bibitem{goodrich2008human}
M.~A. Goodrich and A.~C. Schultz, \emph{Human-robot interaction: a
  survey}.\hskip 1em plus 0.5em minus 0.4em\relax Now Publishers Inc, 2008.

\bibitem{leigh2015person}
A.~Leigh, J.~Pineau, N.~Olmedo, and H.~Zhang, ``Person tracking and following
  with 2d laser scanners,'' in \emph{2015 IEEE International Conference on
  Robotics and Automation (ICRA)}.\hskip 1em plus 0.5em minus 0.4em\relax IEEE,
  2015, pp. 726--733.

\bibitem{kalman1960new}
R.~E. Kalman, ``A new approach to linear filtering and prediction problems,''
  1960.

\bibitem{konstantinova2003study}
P.~Konstantinova, A.~Udvarev, and T.~Semerdjiev, ``A study of a target tracking
  algorithm using global nearest neighbor approach,'' in \emph{Proceedings of
  the International Conference on Computer Systems and Technologies
  (CompSysTech’03)}, 2003, pp. 290--295.

\bibitem{breiman2001random}
L.~Breiman, ``Random forests,'' \emph{Machine learning}, vol.~45, no.~1, pp.
  5--32, 2001.

\bibitem{kuhn1955hungarian}
H.~W. Kuhn, ``The hungarian method for the assignment problem,'' \emph{Naval
  research logistics quarterly}, vol.~2, no. 1-2, pp. 83--97, 1955.

\bibitem{grisetti2007improved}
G.~Grisetti, C.~Stachniss, and W.~Burgard, ``Improved techniques for grid
  mapping with rao-blackwellized particle filters,'' \emph{IEEE transactions on
  Robotics}, vol.~23, no.~1, pp. 34--46, 2007.

\bibitem{hess2016real}
W.~Hess, D.~Kohler, H.~Rapp, and D.~Andor, ``Real-time loop closure in 2d lidar
  slam,'' in \emph{2016 IEEE International Conference on Robotics and
  Automation (ICRA)}.\hskip 1em plus 0.5em minus 0.4em\relax IEEE, 2016, pp.
  1271--1278.

\bibitem{santos2013evaluation}
J.~M. Santos, D.~Portugal, and R.~P. Rocha, ``An evaluation of 2d slam
  techniques available in robot operating system,'' in \emph{2013 IEEE
  International Symposium on Safety, Security, and Rescue Robotics
  (SSRR)}.\hskip 1em plus 0.5em minus 0.4em\relax IEEE, 2013, pp. 1--6.

\bibitem{jolliffe2016principal}
I.~T. Jolliffe and J.~Cadima, ``Principal component analysis: a review and
  recent developments,'' \emph{Philosophical Transactions of the Royal Society
  A: Mathematical, Physical and Engineering Sciences}, vol. 374, no. 2065, p.
  20150202, 2016.

\bibitem{pomerleau2015review}
F.~Pomerleau, F.~Colas, and R.~Siegwart, ``A review of point cloud registration
  algorithms for mobile robotics,'' \emph{Foundations and Trends in Robotics},
  vol.~4, no.~1, pp. 1--104, 2015.

\bibitem{pedregosa2011scikit}
F.~Pedregosa, G.~Varoquaux, A.~Gramfort, V.~Michel, B.~Thirion, O.~Grisel,
  M.~Blondel, P.~Prettenhofer, R.~Weiss, V.~Dubourg \emph{et~al.},
  ``Scikit-learn: Machine learning in python,'' \emph{the Journal of machine
  Learning research}, vol.~12, pp. 2825--2830, 2011.

\end{thebibliography}

\end{document}